\documentclass[letterpaper, 10 pt, conference]{ieeeconf}  

\IEEEoverridecommandlockouts                              

\usepackage{algorithmic}
\usepackage{algorithm}
\usepackage{array}
\usepackage{subfigure}
\usepackage{textcomp}
\usepackage{stfloats}
\usepackage{url}
\usepackage[export]{adjustbox}
\usepackage{verbatim}
\usepackage{graphicx}
\usepackage{cite}
\usepackage{booktabs}
\usepackage{multirow}
\usepackage{siunitx}   
\usepackage{threeparttable}

\usepackage{amsmath}  
\usepackage{amssymb}  

\newcommand{\etal}{\textit{et al.} }

\usepackage{hyperref}
\hypersetup{
    colorlinks=true,
    linkcolor=blue,
    filecolor=blue,      
    urlcolor=blue,
    citecolor=blue,
}

\usepackage{CJKutf8}

\title{ \LARGE \bf TacLoc: Global Tactile Localization on Objects\\ from a Registration Perspective
}

\author{Zirui Zhang, Boyang Zhang, Fumin Zhang and Huan Yin
}

\begin{document}

\maketitle
\thispagestyle{empty}
\pagestyle{empty}

\begin{abstract}
Pose estimation is essential for robotic manipulation, particularly when visual perception is occluded during gripper-object interactions. Existing tactile-based methods generally rely on tactile simulation or pre-trained models, which limits their generalizability and efficiency. In this study, we propose TacLoc, a novel tactile localization framework that formulates the problem as a one-shot point cloud registration task. TacLoc introduces a graph-theoretic partial-to-full registration method, leveraging dense point clouds and surface normals from tactile sensing for efficient and accurate pose estimation. Without requiring rendered data or pre-trained models, TacLoc achieves improved performance through normal-guided graph pruning and a hypothesis-and-verification pipeline. TacLoc is evaluated extensively on the YCB dataset. We further demonstrate TacLoc on real-world objects across two different visual-tactile sensors.
\end{abstract}

\section{Introduction}
Global pose estimation is fundamental to robotic autonomy. For mobile robots, global localization methods enable pose estimation from scratch on a known map~\cite{yin2024survey}. For robotic manipulators, global pose estimation involves determining the pose of the object relative to the robot base, which is critical for ensuring the success of subsequent manipulation tasks. When the end effector is in contact with an object, the object pose can be decomposed as:
\begin{equation}
    \mathbf{T}_{\text{base}}^{\text{obj}}
    = \mathbf{T}_{\text{base}}^{\text{ee}}
    \cdot \mathbf{T}_{\text{ee}}^{\text{obj}}
    \label{eq:pose_decomposition}
\end{equation}
where $\mathbf{T}_{\text{base}}^{\text{obj}} \in \mathrm{SE}(3)$
denotes the object pose in the robot base frame,
$\mathbf{T}_{\text{base}}^{\text{ee}}$ is the end-effector pose
in the base frame, readily available through forward kinematics
of the manipulator, and $\mathbf{T}_{\text{ee}}^{\text{obj}}$
represents the object pose expressed in the end-effector frame.
Since $\mathbf{T}_{\text{base}}^{\text{ee}}$ is known,
the core problem reduces to estimating
$\mathbf{T}_{\text{ee}}^{\text{obj}}$ from sensory measurements
at the end effector.

During manipulation, however, the interactions between the
end effector and the object easily occlude visual perception,
making tactile sensing an increasingly popular modality for
contact-based pose estimation.
Tactile-based global localization thus focuses on estimating
$\mathbf{T}_{\text{ee}}^{\text{obj}}$ from the first touch. Existing studies~\cite{petrovskaya2011global, suresh2023midastouch, bauza2023tac2pose} typically involve rendering tactile data onto the object model and computing the \textit{similarity} with real sensing, which essentially serves as a tactile measurement model for estimating the pose distribution. This distribution is then incorporated into a Monte Carlo localization (MCL) framework~\cite{suresh2023midastouch} or used for precise matching~\cite{bauza2023tac2pose}. However, these methods rely on well-trained networks for codebook building~\cite{suresh2023midastouch} and rendering~\cite{bauza2023tac2pose}, which limits their generalization to different types of tactile sensors or new objects. The efficiency of similarity computation is influenced by the discretization resolution of the $\mathfrak{se}(3)$ space. Furthermore, sequential-based estimation might fail if the finger loses contact with the object.


\begin{figure}[t]
    \centering
    \includegraphics[width=0.95\linewidth]{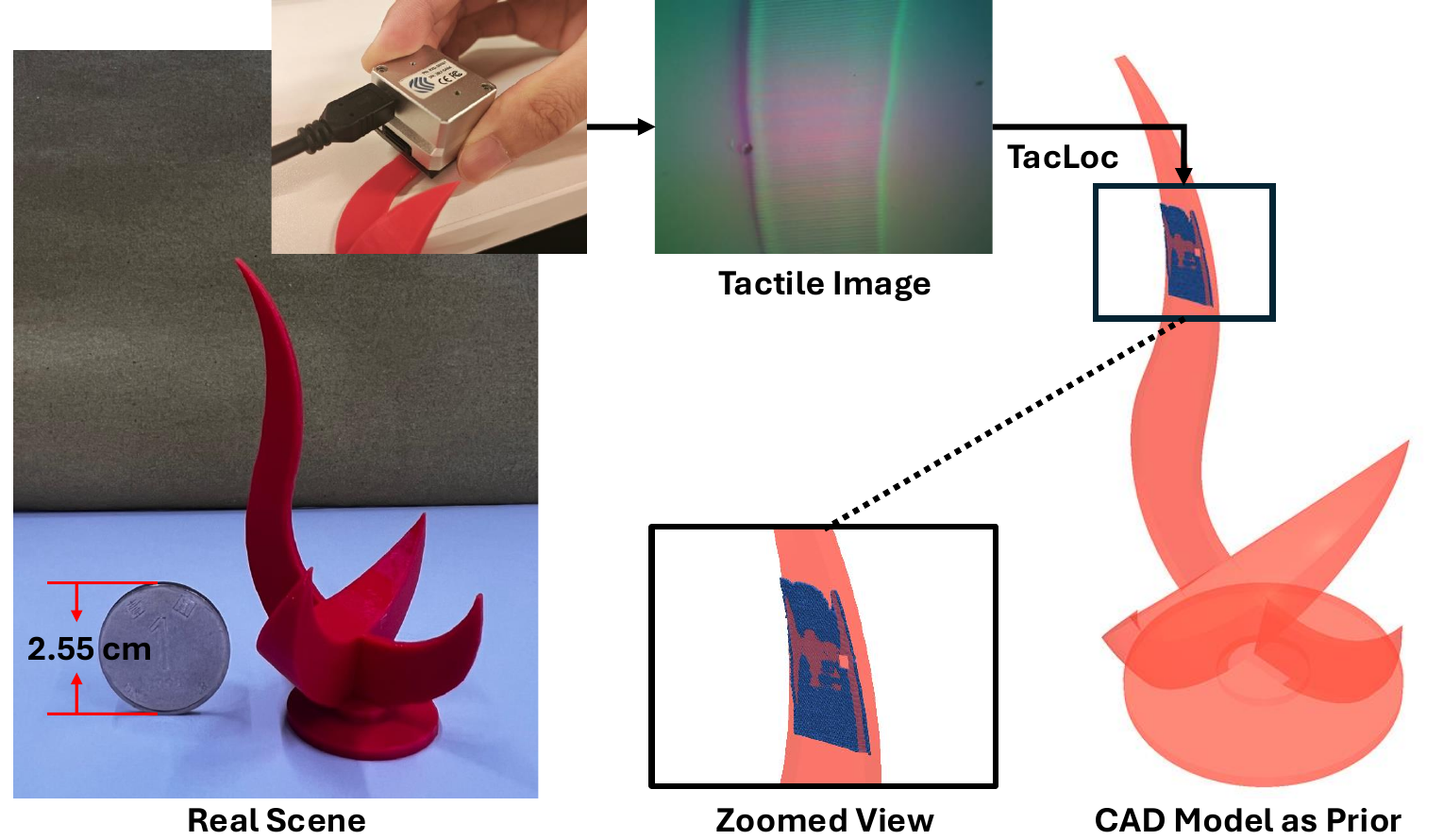}
    \caption{Real-world demonstration using TacLoc. A GelSight Mini sensor touches a textureless 3D-printed object (left), producing a tactile image that is converted into a dense point cloud and registered to a prior CAD model (right). The zoomed view shows the tactile point cloud (blue) aligned on the CAD model (red).}
    \label{fig:teaser}
\end{figure}

One-shot global localization enables direct pose estimation on a given map without sequential filtering in the mobile robotics community~\cite{dube2017segmatch, ma2024tripletloc}, i.e., aligning partial measurements to the full map via global matching, without the similarity comparison. Inspired by this, we aim to answer the question: \textit{Is it feasible to achieve one-shot global localization for contact-based tactile sensing? If so, how can it be done?} Compared to field-level matching for mobile robots, there are two practical challenges for in-hand tactile-based localization: 1) the tactile sensing is fundamentally different from the object model (CAD); 2) deviations between as-designed models and as-is objects introduce challenges in pose estimation. In this study, we propose a novel registration method for global tactile localization, named TacLoc\footnote{Code and data are released at https://anonymous.4open.science/r/Dev-TacLoc-2257.}, to mitigate the impact of the above challenges. Overall, the novelties and contributions can be summarized as follows:
\begin{itemize}
    \item We explore the feasibility of tactile pose estimation from a point cloud registration perspective, different from relying on rendering or pre-trained models as in existing methods.
    \item We design a graph-theoretic partial-to-full registration method for global tactile localization. This reduces the number of edges and computation time by approximately $52\%$ and $93\%$, respectively. 
    
    \item TacLoc is successfully deployed on three different visual-tactile sensors: DIGIT, GelSight, and Daimon. We conduct in-depth analyses on simulated YCB datasets and test TacLoc on 5 real-world household objects, achieving a success rate of 33/50.
\end{itemize}



\section{Related Work}

We first discuss the related work on tactile-based localization. Following this, we review recent trends in point cloud registration.

\subsection{Tactile Pose Estimation} \label{sec:related_tacloc}


Visual-tactile sensors typically provide 2D images on fingers~\cite{GelSight,lambeta2020digit}. Early work~\cite{li2014localization} performed 3 degrees of freedom (DoF) pose estimation by utilizing visual keypoints and applying RANSAC for correspondence estimation, following a typical image matching pipeline. 2D occupancy grid map has also been a basic representation for tactile localization~\cite{pezzementi2011object}.

Robotic manipulation tasks are generally conducted in $\mathrm{SE}(3)$ space, requiring 6 DoF object pose estimation relative to the end effector. Image-to-3D conversion is achieved by recovering depth maps from tactile images~\cite{GelSight}. In the robotics community, MCL has emerged as a powerful Bayesian framework for pose estimation at the back end. Petrovskaya and Khatib~\cite{petrovskaya2011global} incorporated MCL for global tactile localization, demonstrating its applicability on five common objects. Specifically, the object model provides expected contact geometries for building an analytical measurement model, i.e., predicting what the object surface should feel like under various touch configurations. Suresh \etal~\cite{suresh2023midastouch} leveraged a measurement model based on a tactile code network learned from tactile simulation. This model is adapted from LiDAR place recognition~\cite{komorowski2021minkloc3d}. In the work by Bauza \etal~\cite{bauza2023tac2pose}, 2D contact shapes are rendered via simulation and compared with real tactile images to compute the pose distribution. Ota \etal~\cite{ota2023tactile} further enhanced particle filtering with an active action-planning strategy, enabling the robot to minimize observation time while efficiently identifying and localizing the peg part in assembly tasks.

The key aspect of the above works is to render or simulate tactile sensing given the object. Subsequently, the rendered images are fed into convolutional neural networks (CNN)~\cite{bauza2019tactile, bauza2023tac2pose} or point cloud networks~\cite{suresh2023midastouch} for similarity computation. In this study, the proposed TacLoc achieves tactile pose estimation through one-shot global registration, offering a different perspective compared to existing studies.





Recent advances have also introduced multimodal fusion, where visual inputs provide an initial pose, and tactile sensing refines this estimate while complementing the shape of the object~\cite{li2023vihope, vitascope}. While multiple sensors offer richer measurements, it falls outside the scope of our study.


\subsection{Global Point Cloud Registration}

Global registration estimates the relative pose between measured point clouds and a given model. It is typically classified into correspondence-free and correspondence-based approaches~\cite{yin2024survey}. Correspondence-free methods are applicable in pseudo-2D settings, such as the branch-and-bound (BnB) approach in~\cite{3dbbs}. Extending these methods to full 3D scenarios significantly increases computational complexity. For correspondence-based methods, a critical component is the use of local descriptors, like Fast Point Feature Histograms (FPFH)~\cite{fpfh}, to build correspondences. While learning-based descriptors~\cite{ao2021spinnet,poiesi2021distinctive} have shown strong performance, they require large-scale training data and face generalization challenges across different modalities and environments. These challenges are further exacerbated in tactile sensing due to the lack of a well-labeled dataset.

The initial correspondence set derived from descriptors often contains many outliers due to noise or repetitive structures, making effective outlier rejection crucial for accurate transformation estimation. Traditional methods, such as Random Sample Consensus (RANSAC)\cite{fischler1981random}, rely on consensus maximization, but their performance degrades significantly under high outlier ratios. Recent graph-theoretic pruning approaches construct a compatibility graph, where nodes represent correspondences and edges encode mutual consistency. Lusk \etal\cite{clipper} identified the maximum clique for outlier rejection, while Yang \etal~\cite{mac} relaxed the maximum clique to maximal cliques, allowing broader exploration of local consensus for more accurate pose hypotheses. Qiao \etal~\cite{qiao2024g3reg} relaxed the compatibility graph and proposed a distrust-and-verify scheme to select the best hypotheses. This scheme leveraged both sparse features and dense point clouds for the pose verification.

The proposed TacLoc is inspired by existing studies on global registration~\cite{clipper,mac,qiao2024g3reg}. We extract FPFH descriptors from tactile-reconstructed point clouds and design a robust backend for outlier rejection. Specifically, considering the nature of tactile sensing, we introduce a normal-guided graph pruning strategy that enforces sparsity and accelerates the clique search process. The designed pipeline does not require additional training data.


\section{Problem Formulation}\label{sec:problem}

The global localization of mobile robots~\cite{yin2024survey} involves estimating the most probable pose $\mathbf{x} \in  \mathrm{SE}(3)$, given observations $\mathbf{z}$ and a known map $\mathbf{m}$. This estimation is typically formulated within the framework of Bayesian inference as follows:
\begin{equation}
    {\rm p}(\mathbf{x|z,m}) \propto \underbrace{{\rm p}(\mathbf{z|x,m})}_\text{observation} \cdot \underbrace{{\rm p}(\mathbf{x|m})}_\text{prior}
    \label{eq:inference}
\end{equation}

For in-hand tactile localization, $\mathbf{z}$ represents tactile images captured by the sensor, while $\mathbf{m}$ corresponds to the object model that the robot interacts with. For the prior distribution, a common approach involves employing recursive Bayesian filtering~\cite{suresh2023midastouch}. Other approaches~\cite{bauza2023tac2pose, tarf} avoid explicit priors by assuming a uniform distribution over the entire $\mathrm{SE}(3)$ pose space, thus performing exhaustive search. Regarding the measurement model, it can be formulated based on geometric residuals~\cite{petrovskaya2011global} or derived from place recognition results~\cite{suresh2023midastouch}.

Different from the methods above, we model the global localization problem as a \textit{one-shot} localization problem, without the sequential Bayesian inference. Moreover, we formulate a tactile registration pipeline for one-shot pose estimation, guided by a \textit{hypothesis-and-verification }scheme, which proceeds as follows:
\begin{align}
    {\rm p}(\mathbf{x|z,m}) 
    &= \int \underbrace{{\rm p}(\mathbf{x|\boldsymbol\theta,z,m})}_\text{verification} \cdot \underbrace{{\rm p}(\boldsymbol\theta|\mathbf{z,m})}_\text{hypothesis} \, {\rm d}\boldsymbol\theta \notag \\
    &\approx \frac{1}{\sum_{k=1}^Kw_k} \cdot \sum_{k=1}^K w_k \cdot {\rm p}(\boldsymbol\theta_k)
\end{align}
in which we approximate the posterior via $K$ candidate poses $\{\boldsymbol\theta_k\}_{k=1}^K$ and its corresponding weight term $w_k$, i.e., the multiple hypotheses and their confidences. More specifically, the term $\boldsymbol\theta_k$ is a transformation matrix ${\mathbf{T}}_k\in \mathrm{SE}(3)$ in practice; the weight term is estimated in the verification phase. In the proposed TacLoc, the candidate with the maximum likelihood is selected as the final result. More details of the TacLoc are presented in the following section.

\section{Methodology}\label{sec:method}

\begin{figure*}[t]
    \centering
    \includegraphics[width=0.9\linewidth]{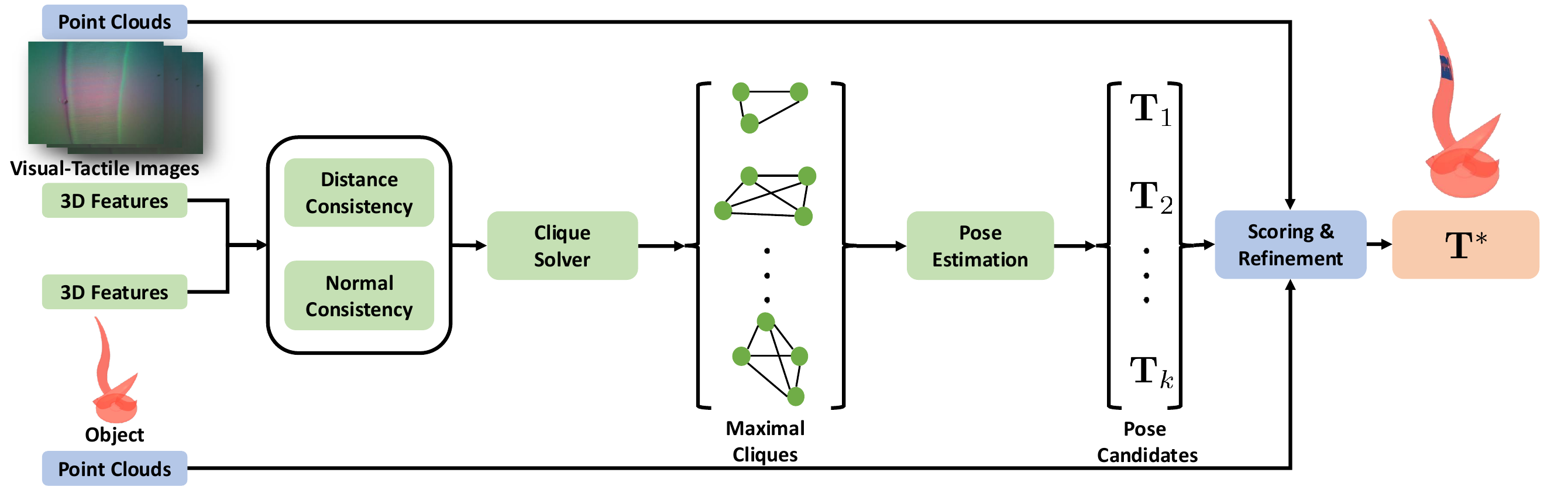}
    \caption{Overview of the TacLoc pipeline for one-shot global tactile localization. The process starts with feature extraction to establish initial correspondences. A compatibility graph is constructed based on distance and normal consistency, and maximal cliques are identified for pose hypothesis generation. Finally, a hypothesis-and-verification approach is applied to refine the pose estimation, achieving partial-to-full tactile localization. It is worth noting that a CAD model is indispensable as a prior map and end-effector poses are required to build submaps during sliding touch.}
    \label{fig:pipeline}
\end{figure*}



We introduce TacLoc by following the pose estimation pipeline, which encompasses both front-end processing and back-end estimation. The front end begins with the raw data obtained from the sensor, while the back end concludes with the pose estimation relative to the object frame.

\subsection{From Raw Data to Initial Correspondence}
\label{sec:preprocessing}

For visual-tactile sensors, raw images are initially processed to recover the height map $\mathbf H$ and gradient maps $\nabla\mathbf{H} = (\mathbf{G}_x, \mathbf{G}_y)$ of the contacted surface. These maps are subsequently converted into point clouds with associated normals to facilitate pose estimation. The method of recovering height maps varies depending on the tactile sensor in use. For instance, the high-resolution DIGIT sensor\cite{lambeta2020digit} directly recovers the height map using a fully convolutional residual network\cite{fcrn}, and gradient maps are derived via partial differentiation of the height map. In contrast, the GelSight Mini (shown in Figure~\ref{fig:teaser}) estimates gradient angle maps $[\tan^{-1}(\mathbf{G}_x), \tan^{-1}(\mathbf{G}_y)]$ using a fully connected neural network that processes five-dimensional features: RGB illumination and XY coordinates. The height map is then computed by solving a Laplacian equation using the discrete cosine transform (DCT)~\cite{GelSight}:
\begin{equation}
    \Delta\mathbf{H} = \frac{\partial^2\mathbf H}{\partial x^2}+\frac{\partial^2\mathbf H}{\partial y^2} = \frac{\partial {\mathbf G}_x}{\partial x}+\frac{\partial {\mathbf G_y}}{\partial y}
\end{equation}
in which $\Delta$ is the Laplacian operator. Generally, the height recovery involves learned regressors provided by the SDK\footnote{For example, height/depth recovery for the GelSight sensor is available at \texttt{https://github.com/GelSightinc/gsrobotics}}.

\begin{figure}[t]
    \centering
    \includegraphics[width=0.95\linewidth]{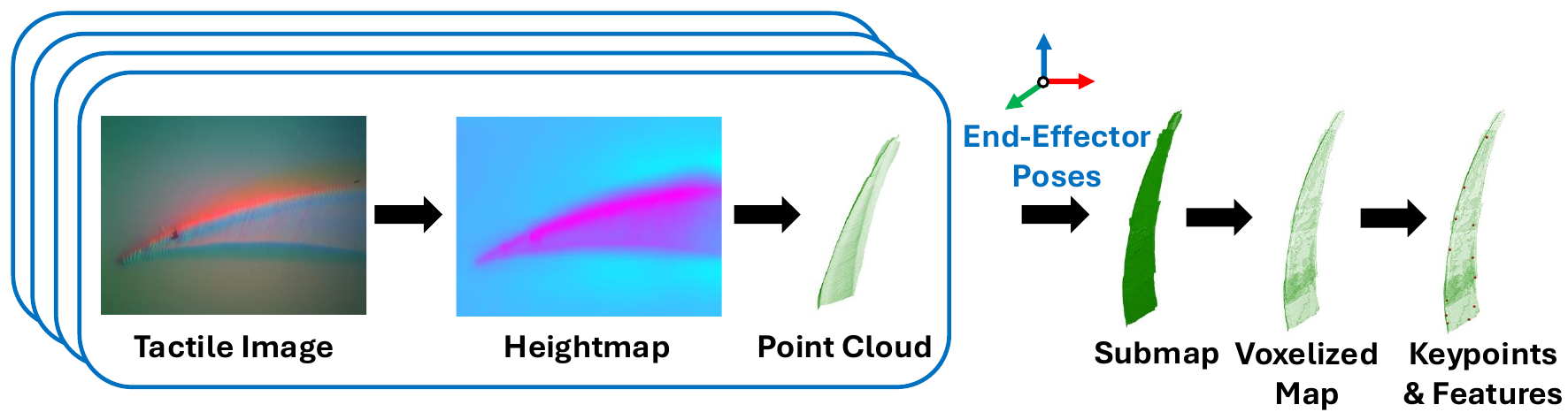}
    \caption{For each frame captured by the tactile sensor, we first convert it into a point cloud with normals. Then we construct a submap representing either a sliding touch or multi-time retouch by integrating end-effector poses. This submap is then processed using our front-end preprocessing pipeline. Best viewed zoomed in and in color.}
    \label{fig:flow}
\end{figure}

The points and their associated normals are derived element-wise using the height map and gradient maps. Given end-effector poses (e.g., obtained through forward kinematics from the manipulator), a point cloud submap can be constructed from a sequence of measurements. Alternatively, a single-shot measurement can also be utilized for one-shot pose estimation. To ensure computational efficiency during pose estimation, voxel grid downsampling is applied to the point clouds. Keypoints are then detected using the Intrinsic Shape Signatures (ISS)~\cite{zhong2009intrinsic} algorithm and are encoded using FPFH~\cite{fpfh} for each keypoint (depicted in Figure~\ref{fig:flow}). Finally, initial correspondences are established by performing Manhattan distance matching in the feature space.

One might argue that learning-based keypoint and descriptor extraction could be a better choice. While we agree that data-driven approaches can improve front-end performance, there are two key considerations: first, there is a lack of publicly available datasets to support learning-based tactile perception, when compared to the extensive datasets available for mobile robots; second, one key contribution of this study is the implementation of the hypothesis-and-verification scheme. We will demonstrate that ISS and FPFH  are simple yet effective within this framework.

\subsection{Multiple Pose Hypotheses Generation}

We first perform graph-theoretic pruning to refine the initial correspondences. Following this, we solve for maximal cliques within the pruned graph to generate multiple pose hypotheses.

\subsubsection{Pairwise Consistency}
Given two corresponded feature points, the feature points from the target and source point clouds are denoted as $\mathcal{P}^\text{tar}=\{\mathbf{p}_i^\text{tar}\in\mathbb{R}^3\}$ and $\mathcal{P}^\text{src}=\{\mathbf{p}_i^\text{src}\in\mathbb{R}^3\}$, respectively. The associated surface normals are denoted as $\mathcal{N}^\text{tar}=\{\mathbf{n}_i^\text{tar}\in\mathbb{R}^3\}$ and $\mathcal{N}^\text{src}=\{\mathbf{n}_i^\text{src}\in\mathbb{R}^3\}$, respectively. We assume that the interacted object is a \textit{rigid body}, meaning that its shape and structure remain unchanged during the manipulation. Consequently, the two correspondences are considered pairwise consistent if they satisfy the following three conditions.

\noindent \textbf{Distance Consistency.} We set the Euclidean difference between two feature points to remain within bounds $\delta_d$ :
\begin{equation}
    \left\vert\lVert\mathbf{p}_i^\text{src} - {\mathbf p}_j^\text{src}\rVert-\lVert{\mathbf p}_i^\text{tar} - {\mathbf p}_j^\text{tar}\rVert\right\vert < \delta_d
\end{equation}
\textbf{Normal Consistency.} Visual-tactile sensing is naturally dense, enabling much more precise surface normal estimation compared to other measurement modalities. We also propose that the absolute angular difference between consistent correspondences lies within a limited range:
\begin{equation}
    \left\vert\angle({\mathbf n}_i^\text{src}, {\mathbf n}_j^\text{src}) - \angle({\mathbf n}_i^\text{tar}, {\mathbf n}_j^\text{tar})\right\vert < \delta_\alpha
\end{equation}
\textbf{Injective Consistency.} Each source point can map to at most one target point, and each target point can also map to at most one source point:
\begin{equation}    \forall \, (i, j), \, i \neq j, \quad \mathbf{p}_i^\text{src} \ne \mathbf{p}_j^\text{src}\land\mathbf{p}_i^\text{tar}\ne\mathbf{p}_j^\text{tar}
\end{equation}
\subsubsection{Maximal Cliques}  
\begin{figure}[t]
    \centering
    \subfigure[Initial Corr.]{
        \includegraphics[width=1.8cm]{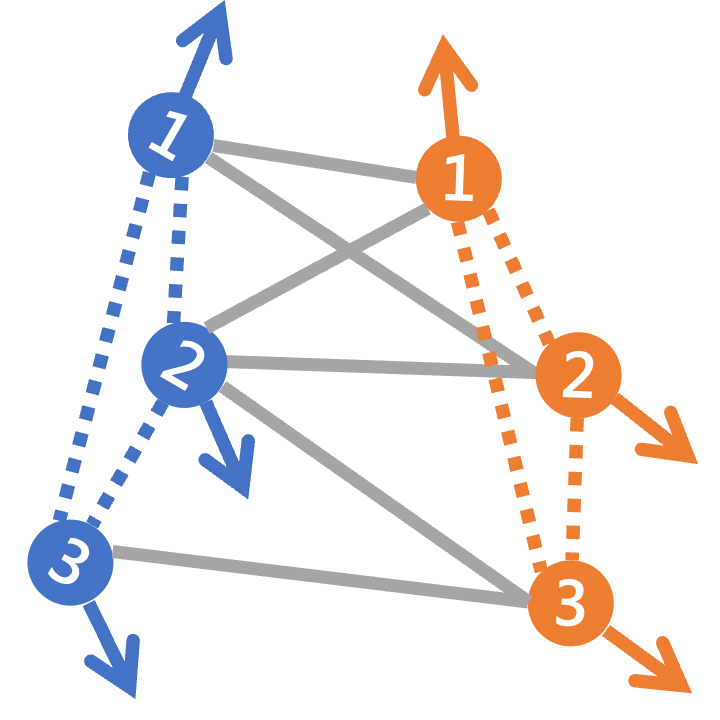}
        \label{fig:correspondence}
    }
    \hfill
    \subfigure[Graph Constr.]{
        \includegraphics[width=1.8cm]{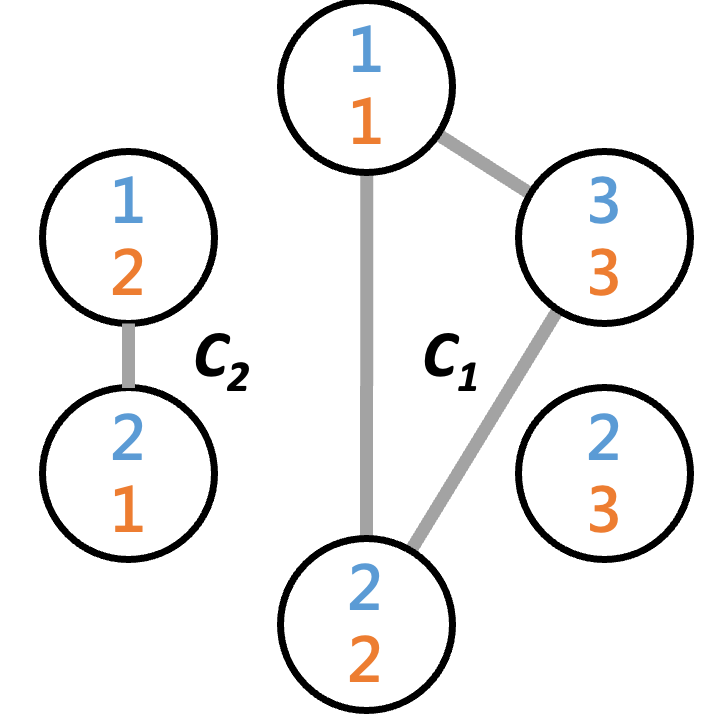}
        \label{fig:graph}
    }
    \hfill
    \subfigure[Clique 1]{
        \includegraphics[width=1.8cm]{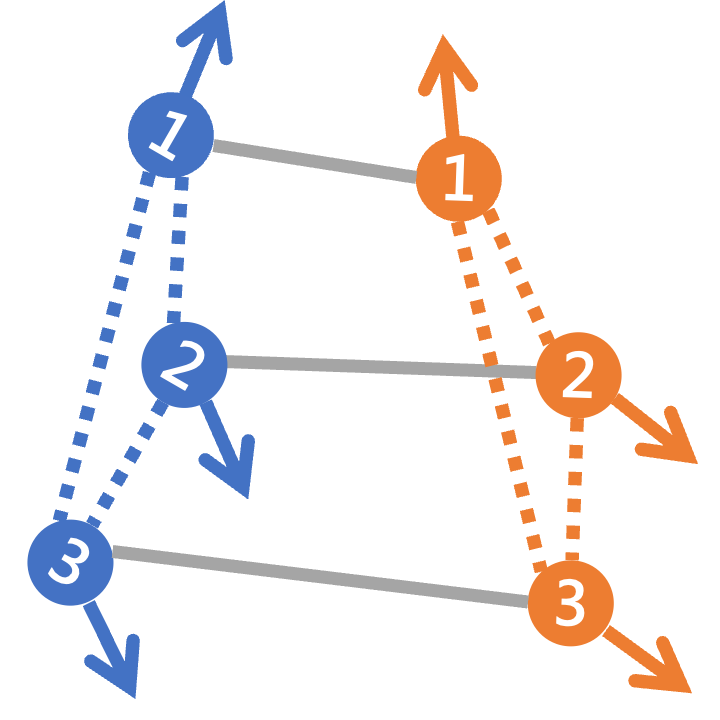}
        \label{fig:clique1}
    }
    \hfill
    \subfigure[Clique 2]{
        \includegraphics[width=1.8cm]{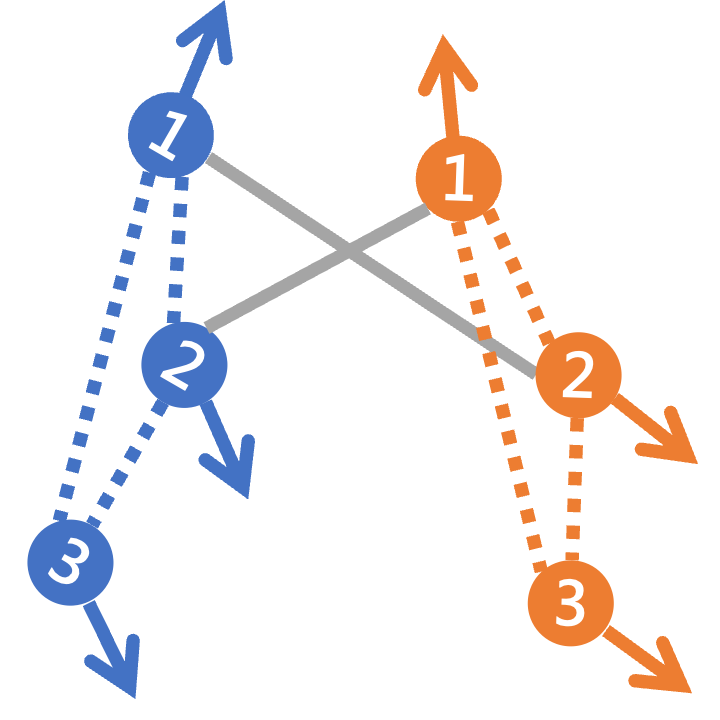}    
        \label{fig:clique2}
    }
    
    \caption{
        Visualization of candidate extraction pipeline with a minimal example (best viewed in color). (a) Three points in the source point cloud (blue) and three points in the target point cloud (orange), connected by six initial correspondences (gray lines). (b) The correspondences form an undirected graph \( G \), with nodes \( V = \{\xi_{1,1}, \xi_{1,2}, \xi_{2,1}, \xi_{2,2}, \xi_{2,3}, \xi_{3,3}\} \) and edges \( E = \{\{\xi_{1,1}, \xi_{2,2}\}, \{\xi_{1,1}, \xi_{3,3}\}, \{\xi_{1,2}, \xi_{2,1}\}, \{\xi_{2,2}, \xi_{3,3}\} \), constructed via pairwise consistency checks. (c) The first maximal clique \( C_1 = \{\xi_{1,1}, \xi_{2,2}, \xi_{3,3}\} \). (d) The second maximal clique \( C_2 = \{\xi_{1,2}, \xi_{2,1}\} \).
    }

        
        
    \label{fig:baby_case}
\end{figure}

A compatibility graph is constructed based on these consistency check criteria above, where nodes represent individual correspondences and edges encode pairwise geometric consistency. Then, maximal cliques are extracted and ranked by size using a modified Bron–Kerbosch algorithm~\cite{bron1973algorithm}. The graph operations and clique extraction are implemented using the NetworkX library \cite{hagberg2008exploring}. A fixed number of top cliques are selected for the following pose estimation and verification.
Figure \ref{fig:baby_case} shows a case study of the candidate extraction pipeline. 

The sparse point clouds reconstructed by range sensing (e.g., laser scanners and radar) often lack sufficient density for consistent normal estimation. In contrast, the dense point clouds obtained from tactile sensors enable robust and accurate normal vector computation. This differs from the distance-only consistency check used in prior methods, such as~\cite{qiao2024g3reg}. It is worth mentioning that 3DMAC~\cite{mac} also incorporates normal consistency for graph pruning through a condition $|\sin\angle({\mathbf n}_i^\text{src}, {\mathbf n}_j^\text{src})-\sin\angle({\mathbf n}_i^\text{tar}, {\mathbf n}_j^\text{tar})|<t_\alpha$. Our approach differs fundamentally and has superiority in time cost. Specifically, while 3DMAC performs post-hoc normal consistency checks, our method conducts preemptive verification, thus ensuring graph sparsity and reducing the complexity of clique extraction.


\subsubsection{Pose Estimation}
\label{sssec:est}


For each selected clique, we estimate its transformation ${\mathbf T}_k=({\mathbf R}_k,{\mathbf t}_k)$ by minimizing both point-to-point and normal-to-normal residuals. The rotation component is estimated as:
\begin{equation} 
\label{eq:residual}
\footnotesize
    \quad {\mathbf R}_k =  \arg\min_{{\mathbf R}_k}\sum_{(i,j)\in C_k}\left( \lVert{\mathbf q}_j'-{\mathbf R}_k{\mathbf p}_i'\rVert^2 + \alpha\cdot\arccos^2\langle{\mathbf m}_j,{\mathbf R}_k{\mathbf n}_i\rangle \right)
\end{equation}
where ${\mathbf p}_i'$ and ${\mathbf q}_j'$ represent centered points from the source and target clouds, respectively; ${\mathbf n}_i$ and ${\mathbf m}_j$ denote their corresponding normals. The term $\alpha$ balances the weight between distance and normal differences; while we observe that the balancing effect is not significant because the two differences are very close. For small angular deviations, where $\cos\langle {\mathbf m}_j, {\mathbf R}_k{\mathbf n}_i \rangle \approx 1$, we utilize the approximation $\arccos^2 \langle {\mathbf m}_j, {\mathbf R}_k{\mathbf n}_i \rangle \approx \lVert {\mathbf m}_j - {\mathbf R}_k{\mathbf n}_i \rVert^2$, based on chordal distance scaling~\cite{chordal_1}. This formulation yields a closed-form solution for Equation~\ref{eq:residual} using the Kabsch Algorithm~\cite{kabsch1978discussion}. Please refer to the cited references for further details.

The translation component is then estimated following~\cite{liu2018efficient}:
\begin{equation}
    {\mathbf t}_k = \arg\min_{{\mathbf t}_k}\sum_{(i,j)\in C_k}\lVert {\mathbf t}_k - ({\mathbf q}_j - {\mathbf R}_k{\mathbf p}_i) \rVert^2
\end{equation}
Finally, we generate $K$ pose candidates based on the pruned correspondences, which are obtained from the maximal cliques used for multiple hypotheses generation.


\subsection{Pose Verification and Refinement}

We utilize a point-to-plane loss function for geometric verification and refinement. This function, which is based on spatial proximity rather than feature descriptors, is expressed as follows:
\begin{equation}
    \mathcal{L}({\delta{\mathbf R}_k}, \delta{\mathbf t}_k)= \sum_i \left\langle 
    \delta {{\mathbf R}_k \hat{\mathbf p}}_i^\text{src} - \tilde{\mathbf{q}}_i + \delta{\mathbf t}_k, \tilde{\mathbf{m}}_i
    \right\rangle^2
\end{equation}
where $\hat{\mathbf{p}}_i^\text{src} = {\mathbf R}_k{\mathbf p}_i^\text{src} + {\mathbf t}_k$, $\tilde{\mathbf q}_i$ denotes the closest point to the transformed source point $\hat{\mathbf{p}}_i^\text{src}$ in the downsampled target cloud, and $\tilde{\mathbf m}_i$ represents the associated normal of $\tilde{\mathbf q}_i$. The refined solution ${\mathbf T}_k^*=(\delta{\mathbf R}_k^*{\mathbf R}_k, \delta{\mathbf R}_k^*{\mathbf t}_k+\delta{\mathbf t}_k^*)$ is regarded as the hypothesis $\boldsymbol\theta_k$ for clique $C_k$. And the weight is granted by $w_k=\exp[-\mathcal{L}(\delta{\mathbf R}_k^*,\delta{\mathbf t}_k^*)]$. The transformations achieving lower residual errors receive higher weights, with the highest-scoring estimate selected as the final pose $\mathbf T^*$, which corresponds to the transformation $\mathbf{T}_{\text{ee}}^{\text{obj}}$ in Equation~\eqref{eq:pose_decomposition}.


It is worth mentioning the generation of target features and points (also referred to as the map in the field of simultaneous localization and mapping (SLAM)). In this study, we sample points based on the CAD model of the objects. Although the modalities of tactile sensing and CAD differ, the sampled points are close to the recovered tactile submaps. This similarity makes one-shot tactile localization feasible. We will validate and evaluate TacLoc in the following section.




\section{Experiments}

We first present the setup and results, analyze robustness and efficiency, and conclude with real-world demonstrations.









\begin{table}[t]
\caption{{Key Parameters of TacLoc for YCB-Reg}}
\label{tab:paras}
\renewcommand{\arraystretch}{1}
\centering
\begin{tabular}{c|c|c}
\hline
\hline
Module & Parameters                     & Value             \\ 
\hline
\multirow{2}{*}{Feature Extraction} & Voxel size               &        $1\text{mm}$       \\
 & Radii               & $5\text{mm}$              \\
\hline
\multirow{2}{*}{Corr. Pruning} & Distance threshold/bound $\delta_d$  & $6\text{mm}$   \\
 & Angular threshold/bound $\delta_\alpha$ & $30^\circ$   \\
\hline
\multirow{2}{*}{Pose Estimation} & Number of pose candidates $K$ & $300$                \\
 & Weight balance $\alpha$ & $1$                \\
\hline
\hline
\end{tabular} \\
\end{table}

\subsection{Set Up and Baselines}

\subsubsection{Datasets}
Inspired by the benchmark in~\cite{suresh2023midastouch}, we select a subset of ten objects from the Yale-CMU-Berkeley (YCB) dataset~\cite{ycb}. Using the high-fidelity simulator TACTO~\cite{tacto}, we simulate sliding motions of 10 cm across the surface of each object using the DIGIT sensor~\cite{lambeta2020digit}. The resulting submaps are generated under a noise-free motion model. Each object is touched ten times, with random starting points, resulting in a total of one hundred motion sequences. We refer to this dataset as the YCB-Reg dataset. The key parameters of TacLoc are shown in Table~\ref{tab:paras}. In addition to the simulation, we also conduct real-world evaluation on real-world objects. The introduction of real-world tests is presented in Section~\ref{sec:realworld}.

\begin{figure}[t]
    \centering
    \newcommand{\benchWidth}{0.17\linewidth}
    \includegraphics[width=\benchWidth]{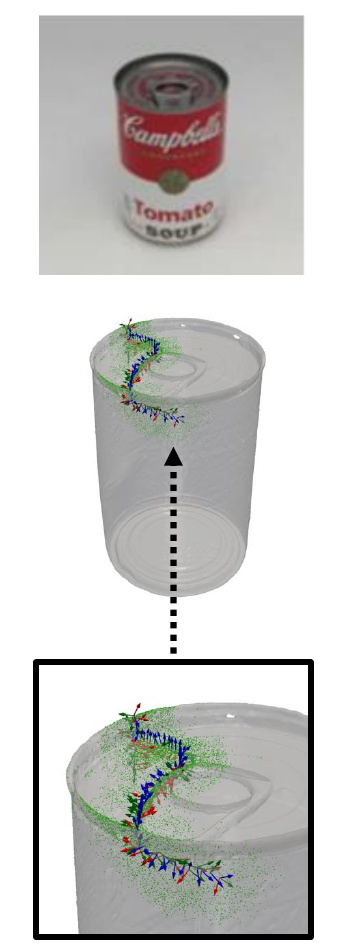}
    \includegraphics[width=\benchWidth]{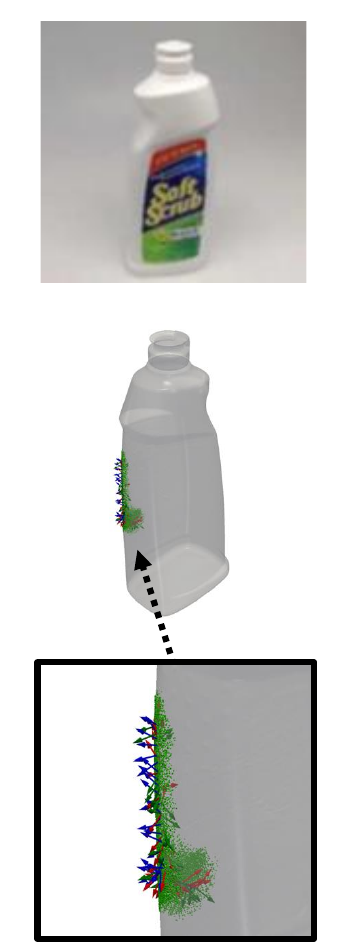}
    \includegraphics[width=\benchWidth]{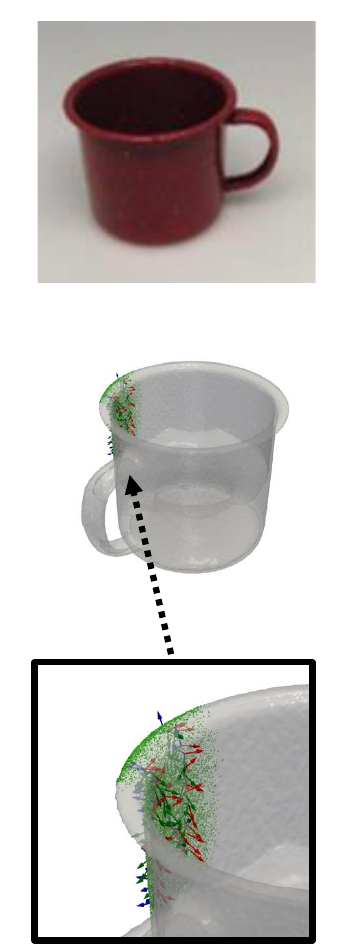} 
    \includegraphics[width=\benchWidth]{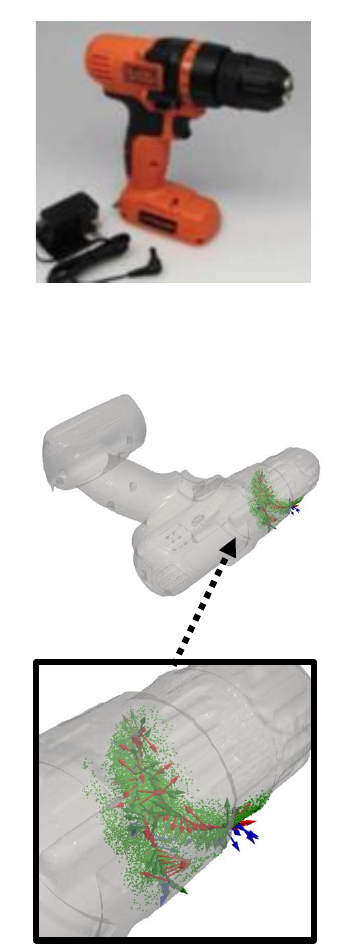}
    \includegraphics[width=\benchWidth]{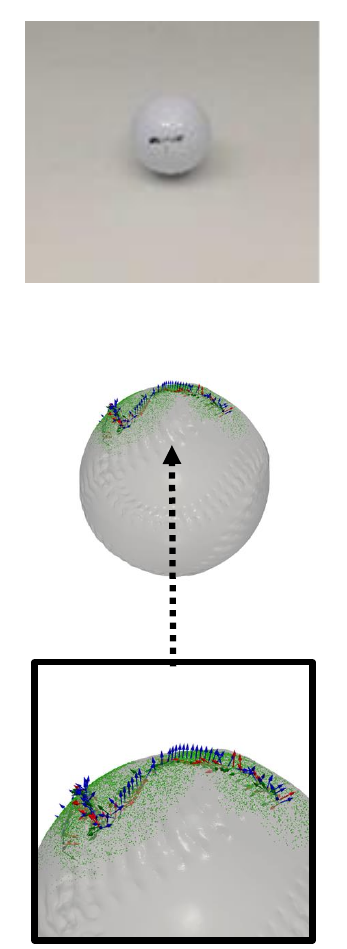}

    \caption{Selected samples from the YCB-Reg benchmark. From left to right are ten objects from the YCB dataset~\cite{ycb}. Gray semi-transparent objects represent the target models, while green point clouds correspond to tactile-based point clouds using the pre-processing approach in Section~\ref{sec:preprocessing}. Zoomed views of these sliding samples are also presented at the bottom.}

\end{figure}

\begin{table}[t]
    \centering
    \caption{Performance Comparison on Tactile Registration}
    \begin{tabular}{c|c|ccc}
        \hline \hline
        Front-end & Back-end & RE (\si{\degree}) & TE (\si{\milli\meter}) & Time (\si{\sec}) \\
        \hline
        FPFH & RANSAC & 128.62 & 99.94 & 1.66 \\
        FPFH & TEASER++ & 19.89 & 8.46 & 13.04 \\
        FPFH & 3DMAC & 19.07 & 9.54 & 2.06 \\
        FPFH & TacLoc & \textbf{0.94} & \textbf{0.69} & \textbf{1.40} \\
        \hline
        SpinNet & RANSAC & 118.63 & 55.89 & $0.05^\star$ \\
        SpinNet & TacLoc & 130.37 & 28.50 & $0.92^\star$ \\
        \hline
        DIP & RANSAC & 137.91 & 66.49 & $0.09^\star$ \\
        DIP & TacLoc & 158.17 & 58.74 & $1.14^\star$ \\
        \hline \hline 
    \end{tabular}
    \begin{tablenotes}
        \item[$\star$] $\star$ \ Time measurements exclude feature extraction.
    \end{tablenotes}
\label{table:comparison}
\end{table}

\subsubsection{Baseline Methods}

We compare TacLoc against state-of-the-art registration methods at both the front end and back end. Specifically, the front-end comparisons involve two advanced learning-based descriptors: SpinNet~\cite{ao2021spinnet} and DIP~\cite{poiesi2021distinctive}. The back-end comparisons include three outlier pruning approaches: RANSAC~\cite{fischler1981random}, TEASER++\cite{teaser}, and 3D MAC\cite{mac}. For front-end baselines, we use the open-source implementations, while for the back-end baselines, exhaustive parameter tuning is performed by varying the inlier thresholds.


        


\subsection{Quantitative Results and Analyses}
\label{sec:quan}


\begin{figure}[t]
    \centering
    \subfigure[Varying Rotation Threshold]{
        \includegraphics[width=4cm]{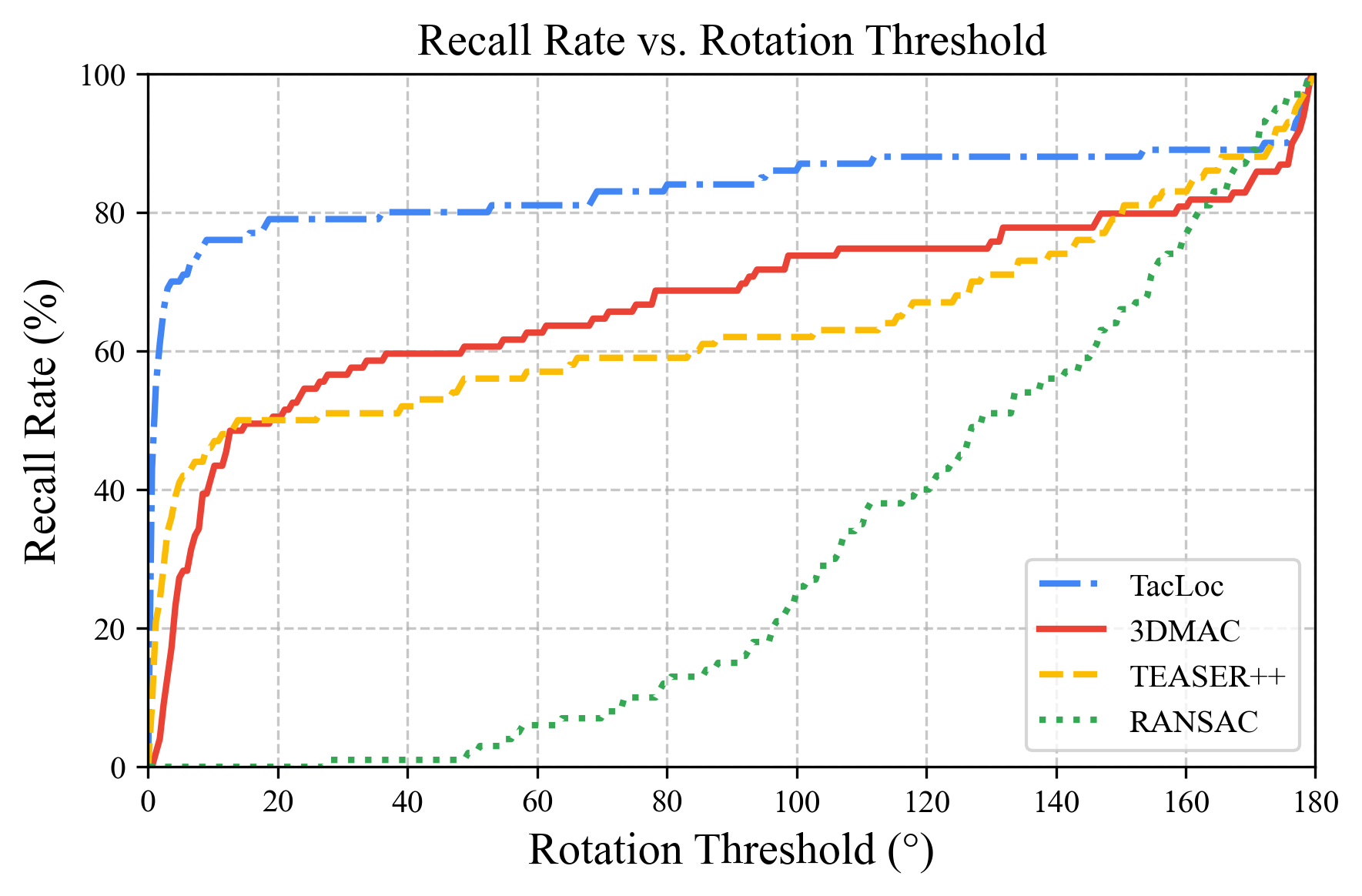}
        \label{fig:cdf_rot}
    }
    \subfigure[Varying Translation Threshold]{
        \includegraphics[width=4cm]{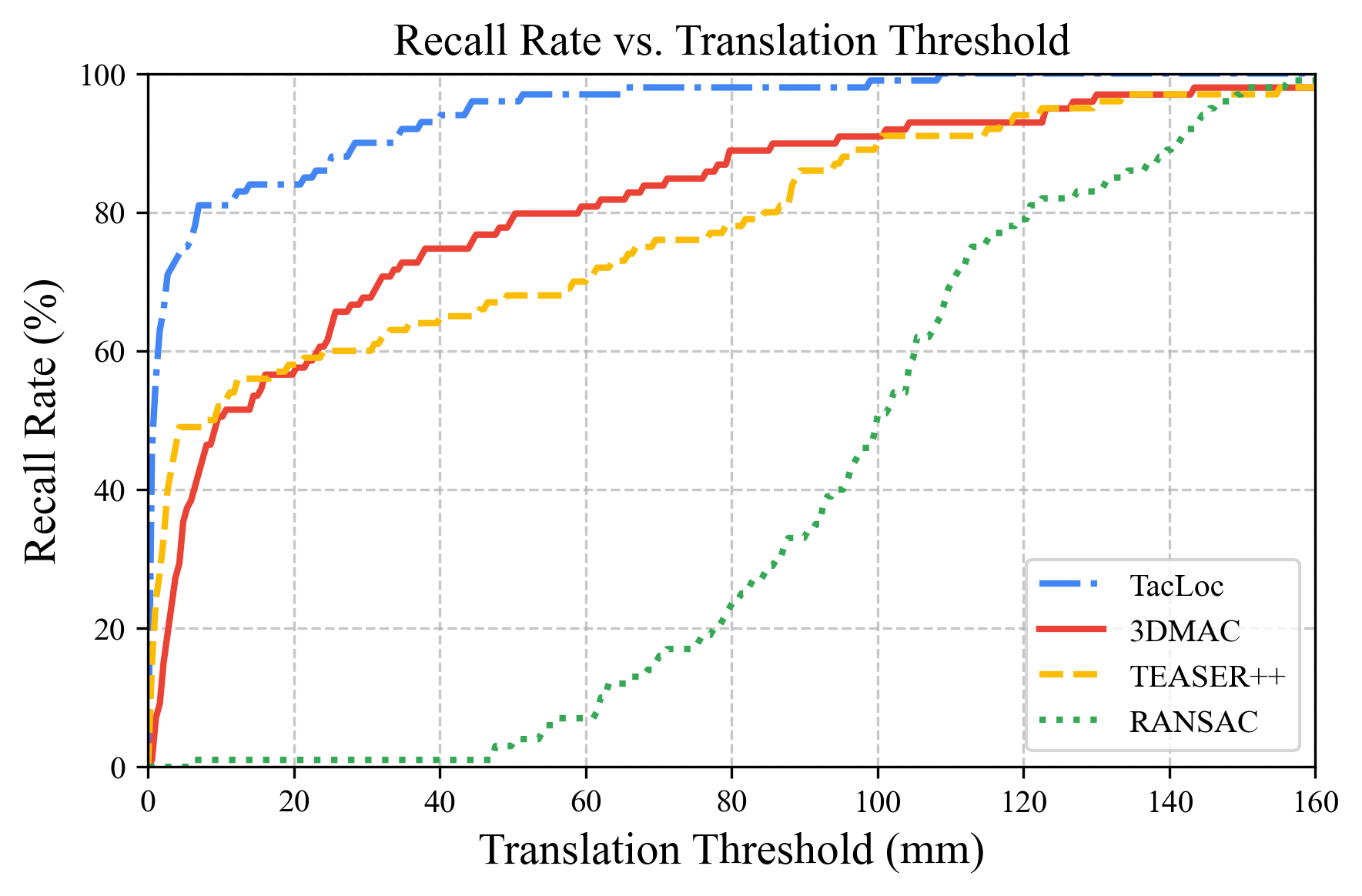}
        \label{fig:cdf_trans}
    }
    \caption{
        Recall rate evaluated on the YCB-Reg using the FPFH descriptor for different back-end approaches under varying criterion of successful registration. The results demonstrate that the proposed TacLoc consistently achieves better performance across all tested conditions.} 
    \label{fig:cdf_results}
\end{figure}

\begin{figure}[t]
    \centering
    \newcommand{\imgWidth}{0.2\textwidth}
    
    \subfigure[\texttt{Adjustable Wrench}]{
        \includegraphics[width=\imgWidth]{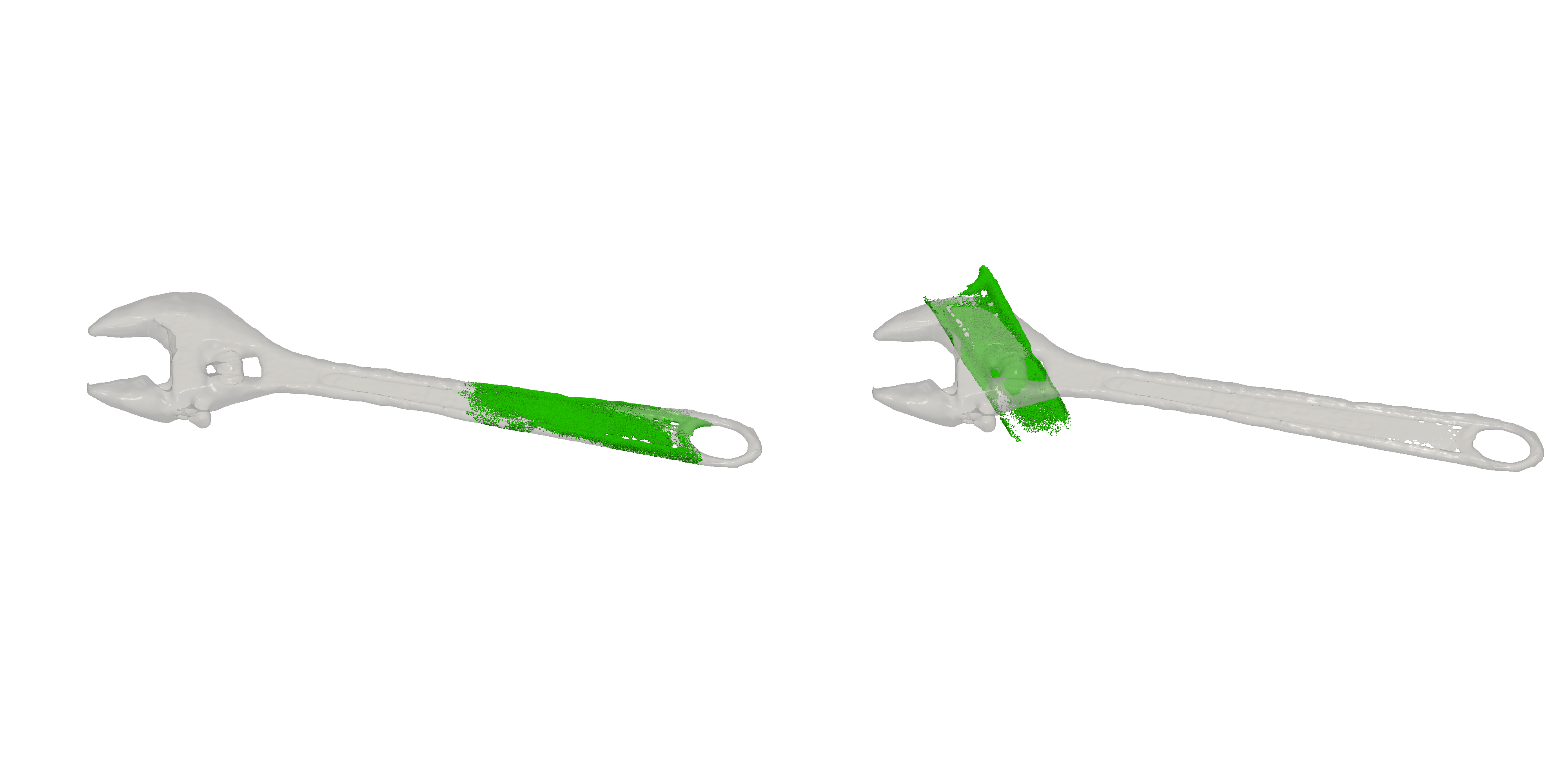}
        \label{fig:series_7}
    }
    \hfill
    \subfigure[\texttt{Sugar Box}]{
        \includegraphics[width=\imgWidth]{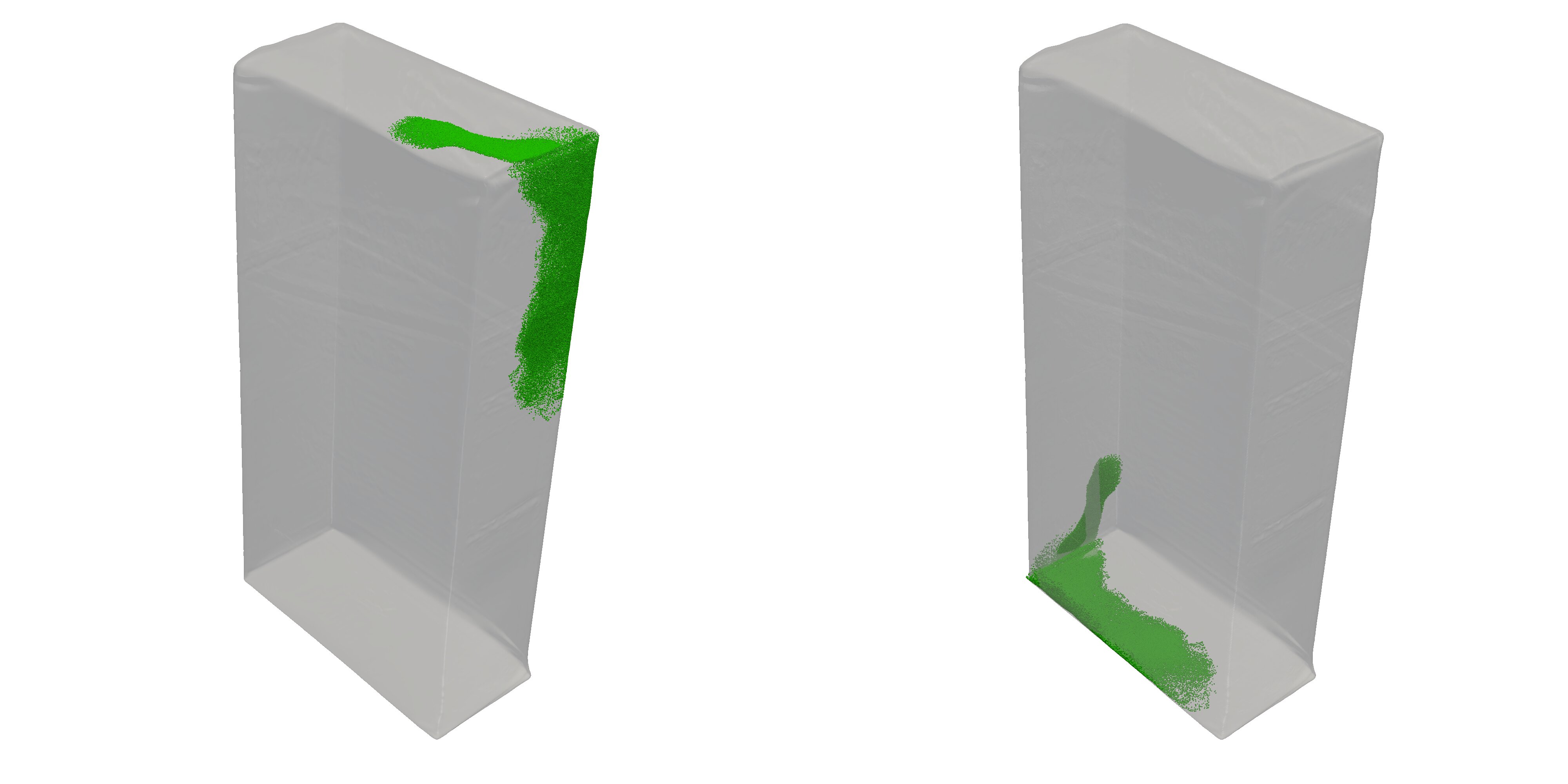}
        \label{fig:series_20}
    }

    \caption{Failure case studies on YCB-Reg benchmark. In each subfigure, the left  shows the ground truth while the right displays the misalignment. TacLoc fails in (a) due to the extremely low inlier ratio; the failure in (b) is caused by repetitive patterns in symmetric objects.
    }
    
    \label{fig:failure_cases}
\end{figure}

The quantitative results on the YCB-Reg benchmark are presented in Table \ref{table:comparison}. With the FPFH front end, TacLoc outperforms other methods across RE and TE. This superior performance can be attributed to two key factors: First, the normal consistency-aided graph effectively maintains graph sparsity while preserving the geometric consistency within each clique, thereby improving the efficiency of clique extraction. Second, the point-to-plane approach, which unifies verification and refinement into a single equation, enhances both rotational and translational accuracy. Additionally, we vary the thresholds for the criterion of successful registration, and present the recall rate in Figure \ref{fig:cdf_results}. The results align consistently with those shown in Table \ref{table:comparison}. We observe that TacLoc does not achieve perfect registration in all cases. From the failure cases, we select two representative examples as case studies, shown in Figure \ref{fig:failure_cases}. 



\subsection{Parameter Sensitivity Analysis}
\label{sec:param}

In addition to the comparisons above, we conduct an in-depth analysis of the proposed TacLoc. Specifically, we evaluate our algorithm on multiple sliding sequences with varying sliding lengths and noise levels in the end-effector pose. For each combination of sliding length and noise level, five sequences are collected using the same procedure as employed in the YCB-Reg benchmark.

\begin{figure}[t]
    \centering
    \subfigure[\texttt{Scissors}]{
        \includegraphics[width=7.4cm]{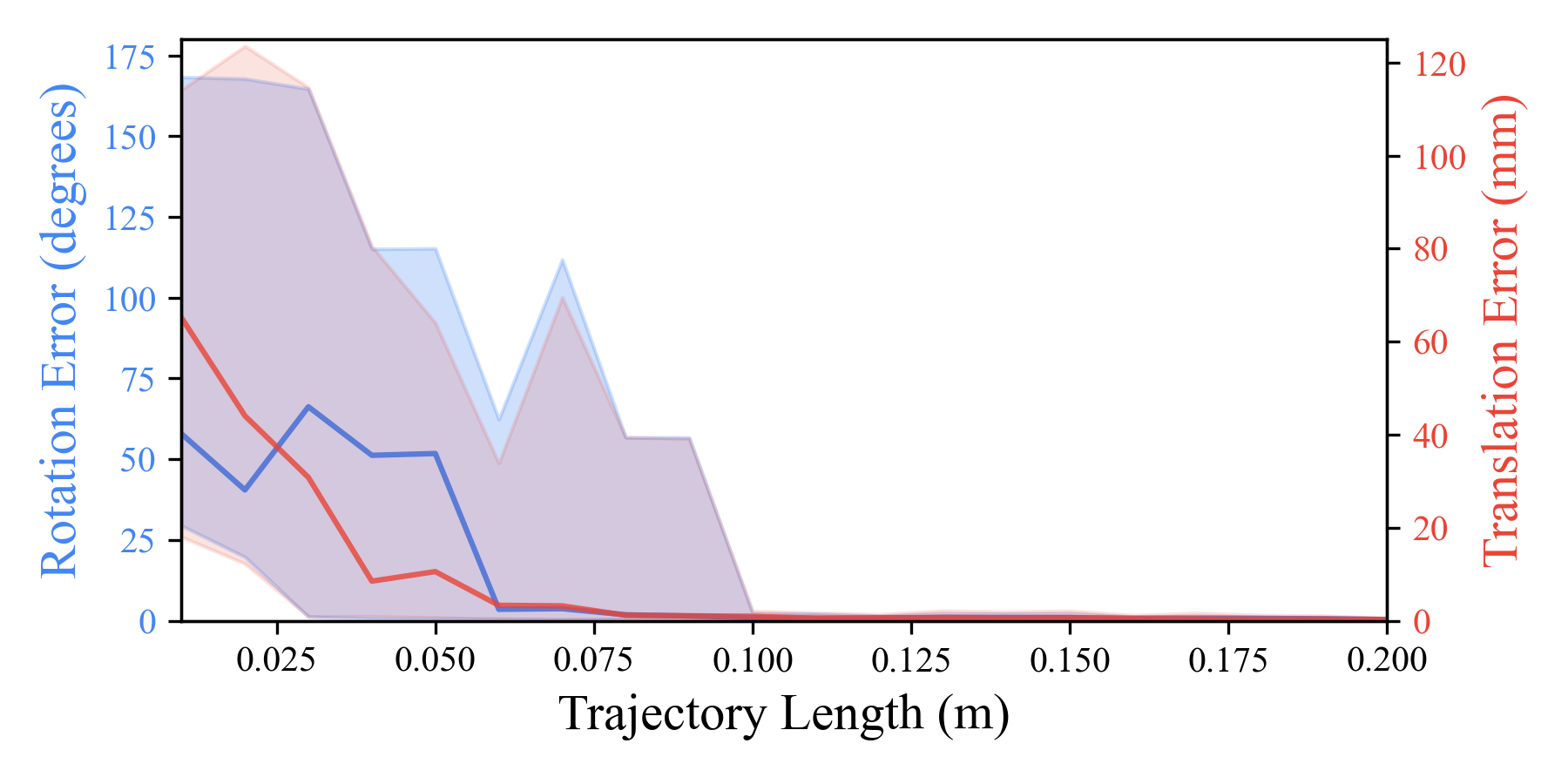}
        \label{fig:ablated:plot}
    }
    \\
    \subfigure[\texttt{Power Drill}]{
        \includegraphics[width=7.0cm]{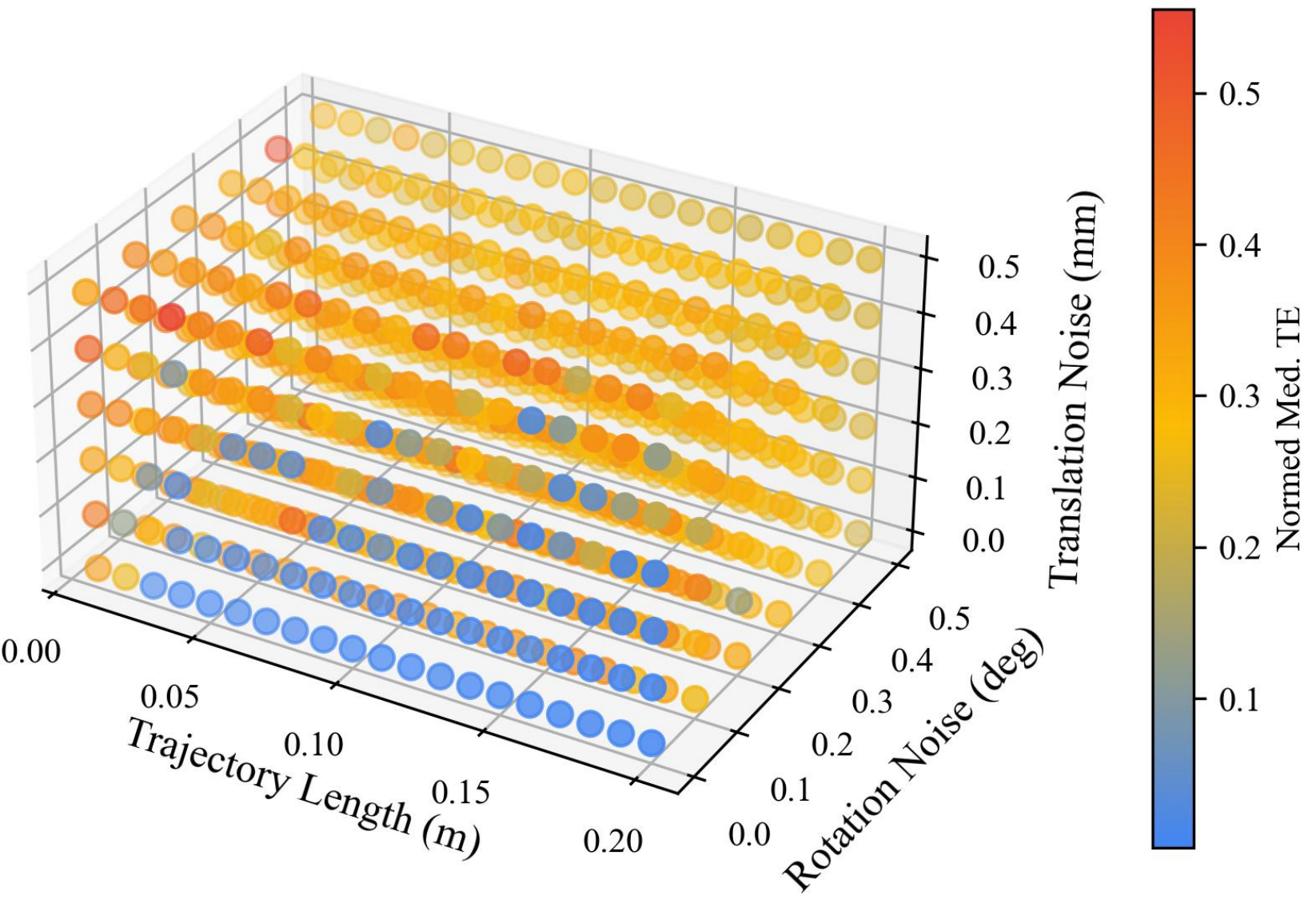}
        \label{fig:ablated:scatter}
    }
    \caption{Parameter analysis on two YCB-Reg objects. A longer sliding length enhances the performance of tactile localization, aligning with human intuition. Additionally, the accuracy of the end-effector pose is critical: a more precise end-effector pose results in a more consistent submap, thereby improving the performance of TacLoc.}
    \vspace{-5pt}
\end{figure}

We first assume a noise-free end-effector pose condition while varying the sliding length on objects. Figure \ref{fig:ablated:plot} presents the median pose error across all five sequences, along with the error range (shaded region). The results demonstrate that the pose error reduces rapidly as the sliding length increases, aligning with the findings reported in MidasTouch~\cite{suresh2023midastouch}. Then we vary the pose noise level and sliding length simultaneously. Figure \ref{fig:ablated:scatter} reports the median translation error, normalized by the object's diagonal length. Intuitively, longer sliding sequences and more accurate end-effector poses lead to more precise results. The results highlight the robustness of TacLoc under conditions with significant noise. On the other hand, the method remains vulnerable to failure when exposed to extreme noise levels.



\subsection{Efficacy of Normal-guided Pruning}

\begin{figure}[t]
    \centering
    \subfigure[Graph sparsification with varying $\delta_\alpha$]{
        \includegraphics[width=8cm]{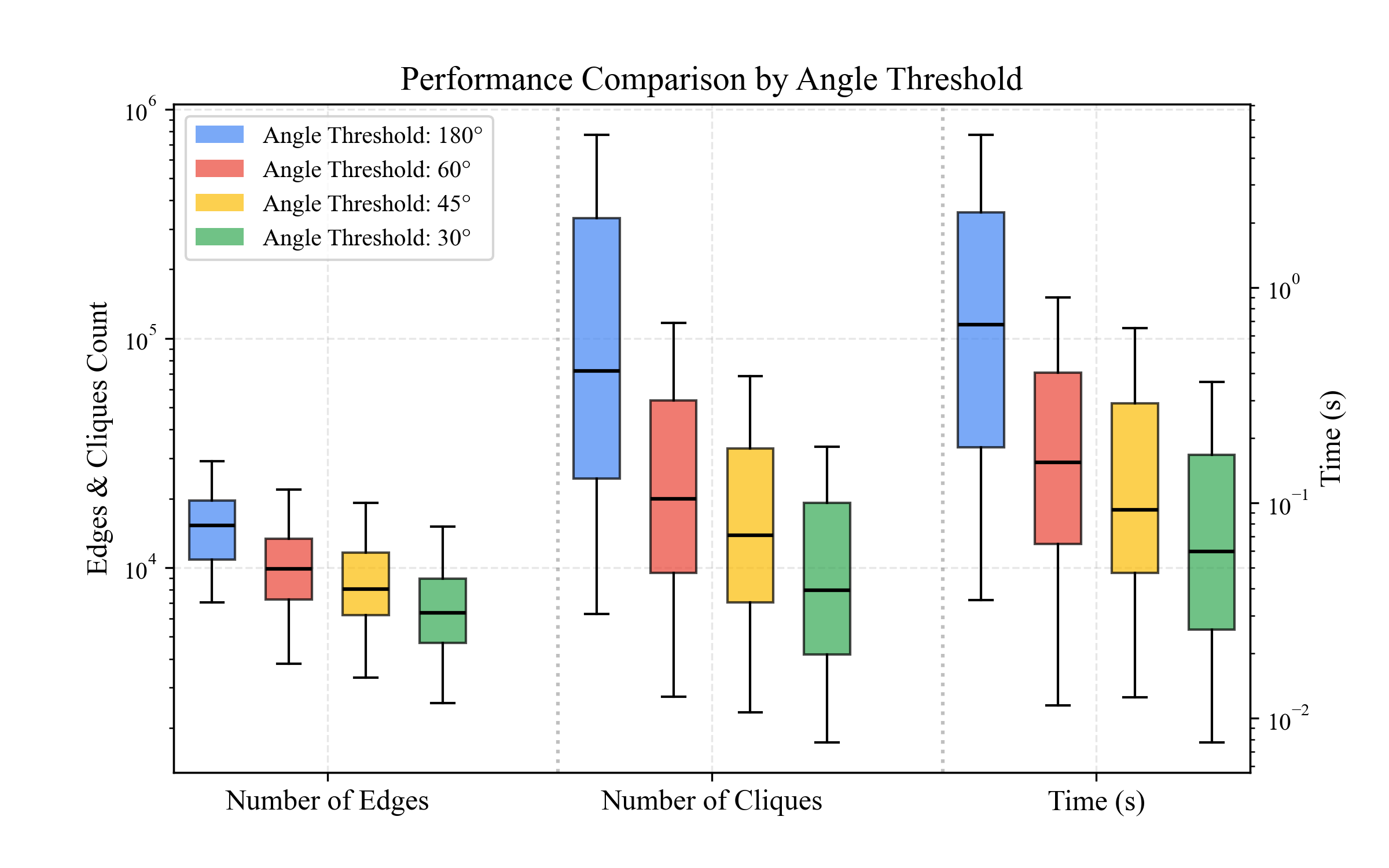}
        \label{fig:complexity:boxplot}
    }
    \hfill
    \subfigure[Scattered points]{
        \includegraphics[width=7.0cm]{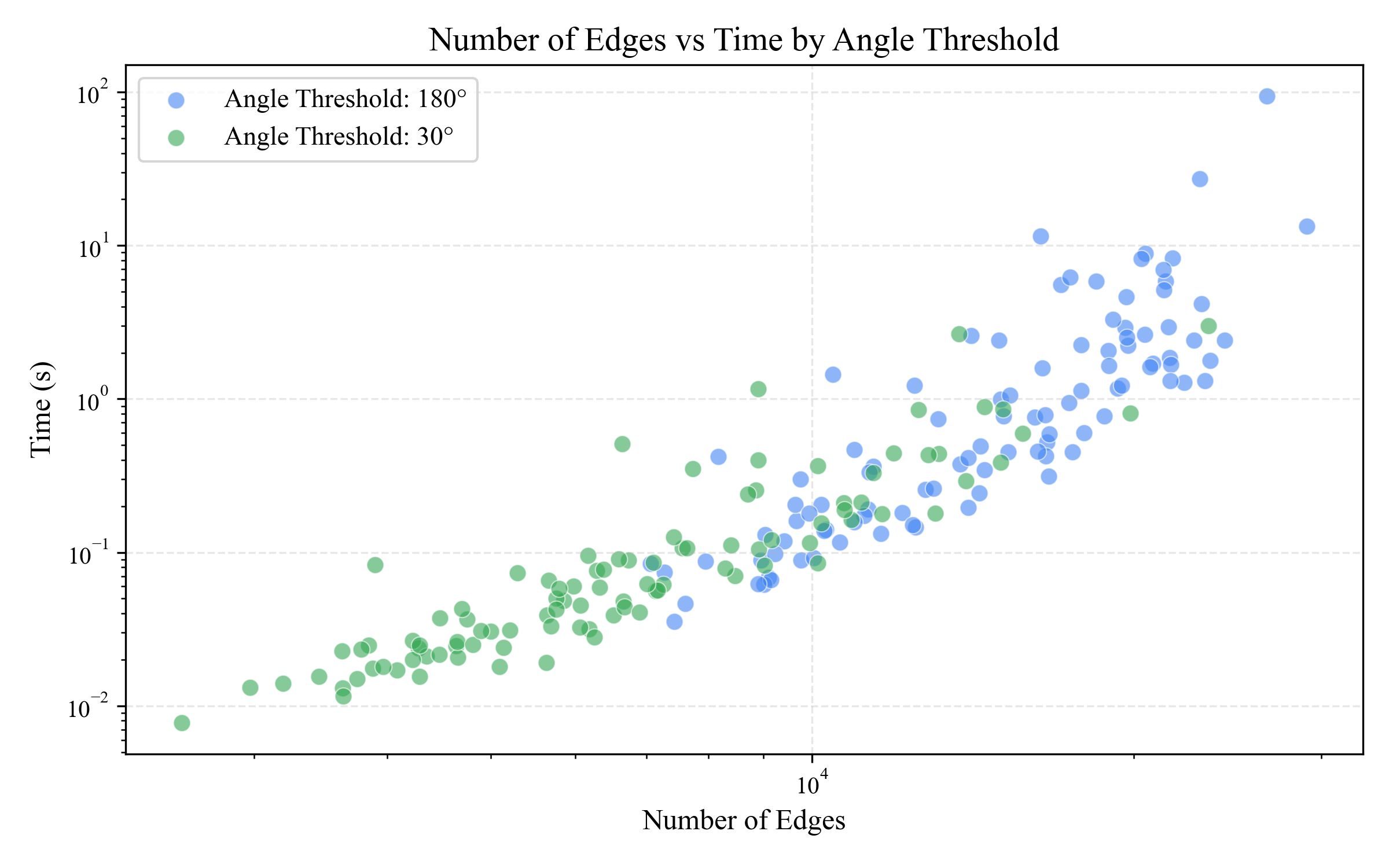}
        \label{fig:complexity:scatter}
    }
    \caption{Experimental results demonstrate the efficacy of normal-guided pruning. (a) Graph density, number of cliques, and computation time decrease as $\delta_\alpha$ varies; (b) The number of edges and computation time exhibit an approximately polynomial relationship. When $\delta_\alpha$ is $30^\circ$ or $180^\circ$, the average number of edges and computation times are $7,473.82$ edges with $0.20\text{s}$ and $15,476.75$ edges with $2.87\text{s}$, respectively.}
    \label{fig:complexity:plot}
    \vspace{-5pt}
\end{figure}


To validate the efficiency improvements achieved through normal-guided pruning, we adjust the normal consistency threshold $\delta_\alpha$ across four different values ranging from $180^\circ$ to $30^\circ$. For each threshold setting, we conduct 100 registration trials on the YCB-Reg dataset while maintaining identical front-end processing pipeline that consistently selected the same 500 correspondences to construct the consistency graph. As shown in Figure \ref{fig:complexity:boxplot}, tightening $\delta_\alpha$ consistently resulted in significant reductions in both edge density and the number of maximal cliques, leading to a pronounced decrease in computation time.

We further visualize the data distributions for $\delta_\alpha = 180^\circ$ and $\delta_\alpha = 30^\circ$ in Figure \ref{fig:complexity:scatter}. Theoretically, maximal clique search has an exponential worst-case complexity of $O(3^{n/3})$ with respect to the number of graph nodes~\cite{tomita2006worst}. On the other hand, after applying normal-guided pruning, our results demonstrate an approximately polynomial relationship between the number of edges and computational cost. These findings indicate that normal-guided pruning effectively reduces the computational complexity of maximal clique search, thereby improving the efficiency of tactile localization.


\subsection{Computational Efficiency Analysis}

\begin{figure}[t]
    \centering
    \includegraphics[width=0.85\linewidth]{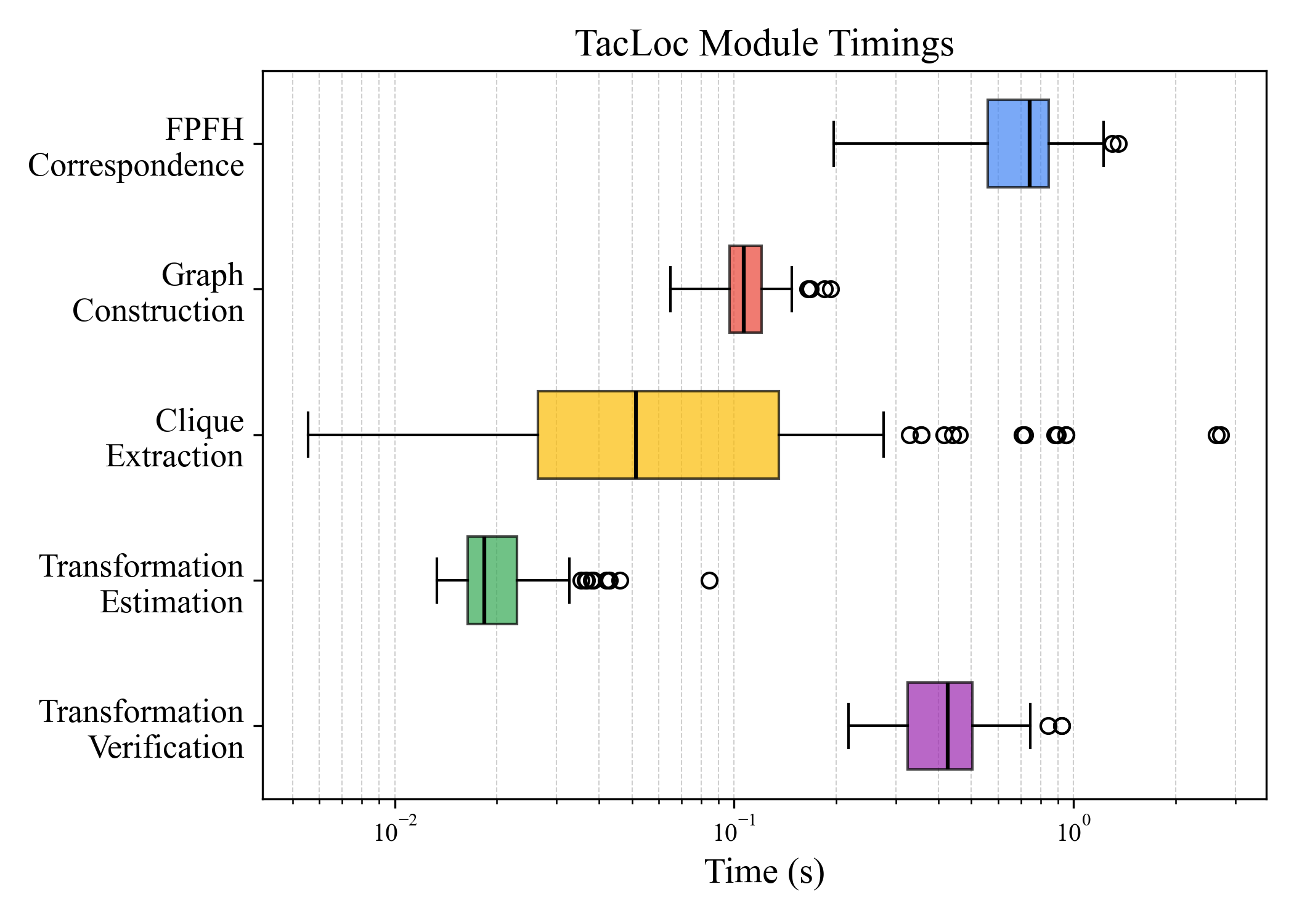}
    \caption{Per-stage computational profile. FPFH-based initial data association and transformation verification are the most time-intensive stages. }
    \label{fig:timings}
\end{figure}

We conduct a detailed per-stage computational profile to analyze the computational efficiency of TacLoc. All timing experiments were performed under CPU-serialized conditions on a portable computing device equipped with an Intel processor and 32GB of RAM. Figure~\ref{fig:timings} provides a statistical breakdown of time consumption across the different stages of TacLoc, revealing that FPFH-based initial correspondence and transformation verification are the most time-intensive stages.

\begin{figure}[t]
    \centering
    \subfigure[$\delta_d$]{
        \includegraphics[width=0.45\linewidth]{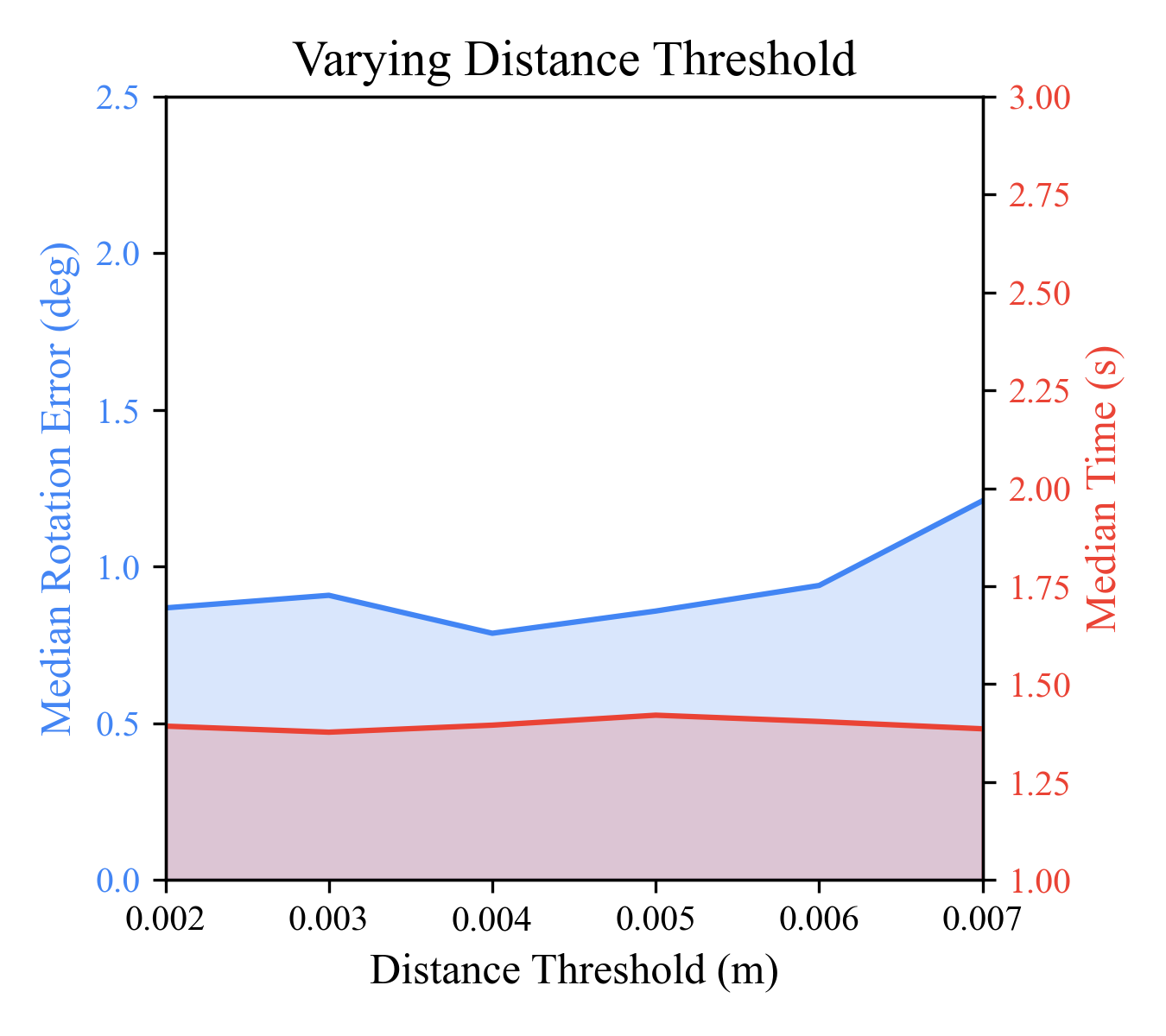}
    }
    \subfigure[$\delta_\alpha$]{
        \includegraphics[width=0.45\linewidth]{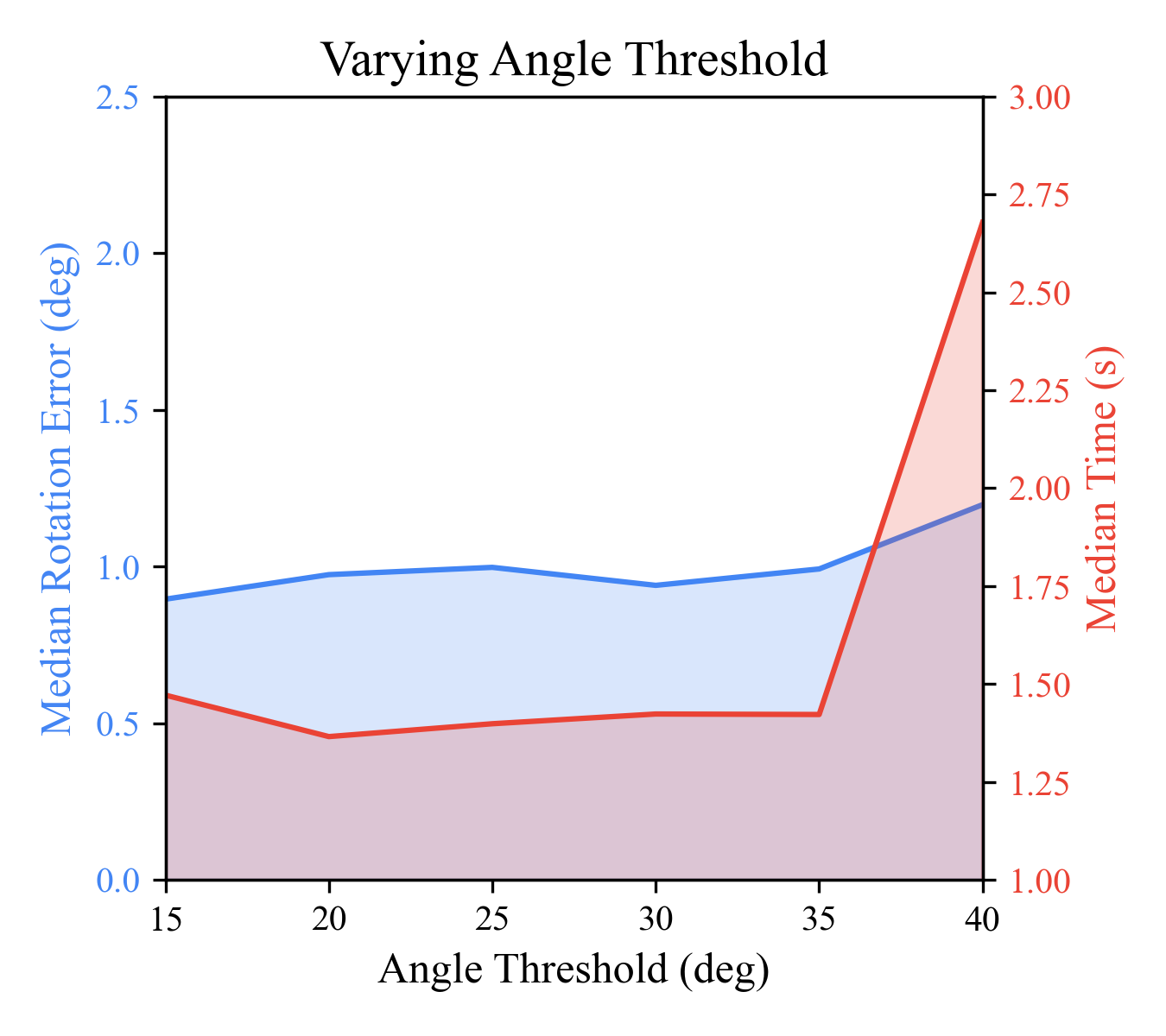}
    }
    \caption{Rotation error and time cost with varying thresholds $\delta_d$ and $\delta_\alpha$.}
    \label{fig:plot}
\end{figure}

In addition, we evaluate the efficiency and effectiveness of the method under varying $\delta_d$ and $\delta_\alpha$ in the consistency check. Theoretically, larger threshold values result in denser graph constructions, leading to longer computation times, while an increased number of incorrect correspondences introduces larger estimation errors. This analysis aligns with the conclusions drawn from Figure~\ref{fig:plot}. 




\begin{figure}[t]
    \centering
    \subfigure[]{
    \includegraphics[width=0.45\linewidth]
    {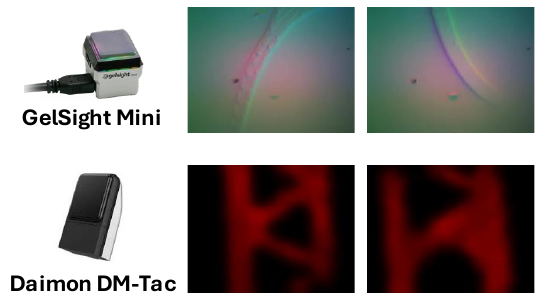}
    }
    \subfigure[]{
    \includegraphics[width=0.45\linewidth]
    {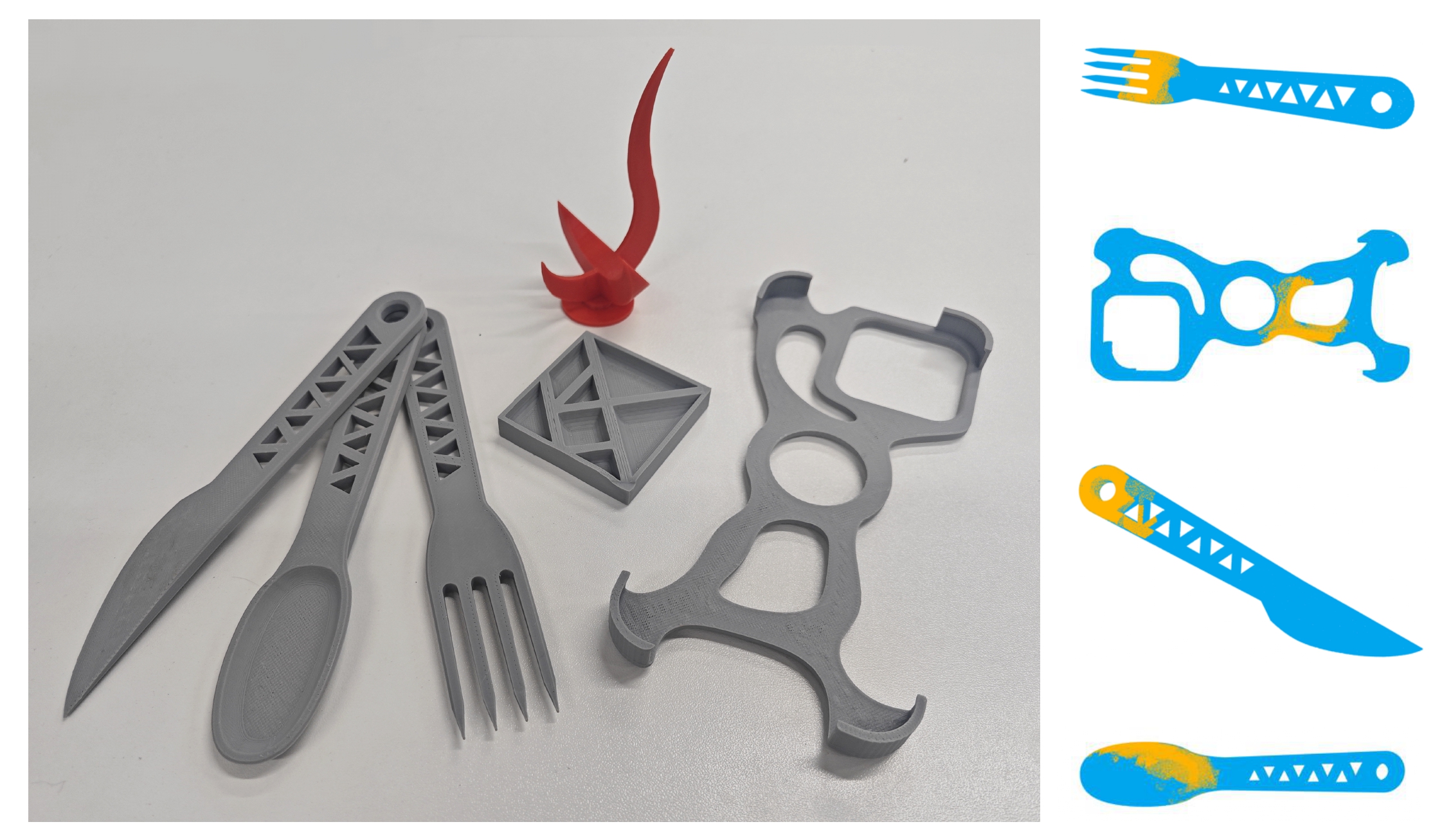}
    \label{fig:real world}
    }
    \caption{(a) TacLoc is deployed on two real-world sensors beyond the DIGIT used in YCB simulation, despite these sensors having different characteristics. The right side shows the visual-tactile images obtained from the sensors. (b) Evaluation on five real-world objects using the GelSight Mini. We perform sliding touches on the objects 50 times, and the proposed TacLoc achieves a success rate of 33/50. Note that the red object is used for the single-touch demonstration shown in Figure~\ref{fig:teaser}. The right side displays the successful cases, where the yellow points represent the submaps.}
    \label{fig:sensors}
\end{figure}

\subsection{Evaluation on Real-world Objects}
\label{sec:realworld}

To validate the generalization beyond simulation, we conduct exploratory tests on household objects using the GelSight Mini sensor. As shown in Figure~\ref{fig:real world}, we deploy TacLoc on a knife, spoon, fork, tangram, and phone case\footnote{The CAD models used can be found at \texttt{https://makerworld.com}}. ICP~\cite{besl1992method} or NormalFlow~\cite{huang2024normalflow} is used to obtain end-effector poses (forward kinematics is optional if a robot arm is used). Obtaining ground truth for tactile localization in the real world is challenging, so we consider the method to achieve successful localization if it produces visually plausible registrations. The overall success rate is 33/50 for the sliding touches on these five objects. We observe a higher success rate when the contact regions contain sufficient geometric features, with particularly clear alignments on objects exhibiting distinctive curvatures.

One might ask how partial or incorrect CAD models impact performance. When touching missing or incorrect parts, deviations between the actual tactile sensing and the designed CAD models will lead to incorrect correspondences and localization failures. Therefore, accurate CAD models are critical for tactile pose estimation. On the other hand, the proposed TacLoc could achieve tactile localization, though manufacturing defects exist in Figure~\ref{fig:real world}.

Until now, we have tested DIGIT~\cite{lambeta2020digit} in simulation and GelSight~\cite{GelSight} on real-world objects. Additionally, we explore deploying TacLoc using a new Daimon sensor (Figure~\ref{fig:sensors}). We carefully tune the sensing component (like the scale) to obtain point clouds and successfully deploy TacLoc to achieve global tactile localization. These sensors exhibit different resolution characteristics, confirming the fundamental compatibility of proposed method across various sensing technologies. Notably, we observe that varying noise profiles and spatial resolutions can influence correspondence density, suggesting that sensor-specific tuning could further enhance performance.



\section{Conclusion}


In this study, we design a normal-aided graph pruning-based point cloud registration method for global tactile localization. We analyze and evaluate the proposed method in both simulation and real-world experiments. We consider two promising directions for future work. First, the fusion of tactile data from multiple tactile measurements~\cite{yu2025multi} could be explored to enhance the accuracy and robustness of pose estimation through collaborative perception. Second, investigating strategies for active tactile exploration~\cite{ota2023tactile} could enable robots to interact with specific object regions while refining pose estimation.






\bibliographystyle{IEEEtran}
\bibliography{main}

\end{document}